# Collision detection or nearest-neighbor search? On the computational bottleneck in sampling-based motion planning[*]

Michal Kleinbort[1][**], Oren Salzman[2][**], and Dan Halperin[1]

[1] Blavatnik School of Computer Science, Tel-Aviv University, Israel
[2] Carnegie Mellon University, Pittsburgh PA 15213, USA

**Abstract.** The complexity of nearest-neighbor search dominates the asymptotic running time of many sampling-based motion-planning algorithms. However, collision detection is often considered to be the computational bottleneck in practice. Examining various asymptotically optimal planning algorithms, we characterize settings, which we call *NN-sensitive*, in which the *practical* computational role of nearest-neighbor search is far from being negligible, i.e., the portion of running time taken up by nearest-neighbor search is comparable, or sometimes even greater than the portion of time taken up by collision detection. This reinforces and substantiates the claim that motion-planning algorithms could significantly benefit from efficient and possibly specifically-tailored nearest-neighbor data structures. The asymptotic (near) optimality of these algorithms relies on a prescribed connection radius, defining a ball around a configuration $q$, such that $q$ needs to be connected to all other configurations in that ball. To facilitate our study, we show how to adapt this radius to non-Euclidean spaces, which are prevalent in motion planning. This technical result is of independent interest, as it enables to compare the radial-connection approach with the common alternative, namely, connecting each configuration to its $k$ nearest neighbors ($k$-NN). Indeed, as we demonstrate, there are scenarios where using the radial connection scheme, a solution path of a specific cost is produced ten-fold (and more) faster than with $k$-NN.

## 1 Introduction

Given a robot $\mathcal{R}$ moving in a workspace $\mathcal{W}$ cluttered with obstacles, motion-planning (MP) algorithms are used to efficiently plan a path for $\mathcal{R}$, while avoiding collision with obstacles [13, 36]. Prevalent algorithms abstract $\mathcal{R}$ as a point in a high-dimensional space called the *configuration space* (C-space) $\mathcal{X}$ and plan a path (curve) in this space. A point, or a configuration, in $\mathcal{X}$ represents a placement of $\mathcal{R}$ that is either collision-free or not, subdividing $\mathcal{X}$ into the sets $\mathcal{X}_{\text{free}}$ and $\mathcal{X}_{\text{forb}}$, respectively. *Sampling-based* algorithms study the structure of $\mathcal{X}$ by constructing a graph, called a *roadmap*, which approximates the connectivity of $\mathcal{X}_{\text{free}}$. The nodes of the graph are collision-free configurations sampled at random. Two (nearby) nodes are connected by an edge if the straight line segment connecting their configurations is collision-free as well.

---

[*] This work has been supported in part by the Israel Science Foundation (grant no. 825/15), by the Blavatnik Computer Science Research Fund, and by the Hermann Minkowski–Minerva Center for Geometry at Tel Aviv University. O. Salzman has been also supported by the German-Israeli Foundation (grant no. 1150-82.6/2011), by the National Science Foundation IIS (#1409003), Toyota Motor Engineering & Manufacturing (TEMA), and the Office of Naval Research. Parts of this work were done while O. Salzman was a student at Tel Aviv University.
[**] M. Kleinbort and O. Salzman contributed equally to this paper.

Sampling-based MP algorithms are typically implemented using two primitive operations: *Collision detection* (CD) [39], which is primarily used to determine whether a configuration is collision-free or not, and *Nearest-neighbor* (NN) search, which is used to efficiently return the nearest neighbor (or neighbors) of a given configuration. CD is also used to test if the straight line segment connecting two configurations lies in $\mathcal{X}_{\text{free}}$—a procedure referred to as *local planning* (LP). In this paper we consider both CD and LP calls when measuring the time spent on collision-detection operations.

*Contribution* The complexity of NN search dominates the asymptotic running time of many sampling-based MP algorithms However, the main computational bottleneck in practical settings is typically considered to be LP [13, 36]. In this paper we argue that this may not always be the case. We describe settings, which we call *NN-sensitive*, where the (computational) role of NN search after finite running-time is far from negligible and merits the use of advanced and specially-tailored data structures; see Fig. 1 for a plot demonstrating this behavior. NN-sensitive settings may be due to (i) planners that *algorithmically* shift the computational weight to NN search; (ii) scenarios in which certain planners perform mostly NN search; or (iii) parameters' values for which certain planners spend the same order of running time on NN and CD.

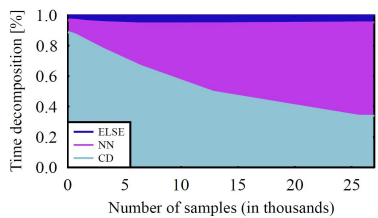

**Fig. 1:** Running-time breakdown of the main primitive operations used in MPLB [46] applied to the 3D-Grid scenario (Fig. 2a). For additional data, see Sec. 4. Best viewed in color.

Specifically, we focus on asymptotically (near) optimal MP algorithms. These planners, which are the standard practice nowadays, are of specific interest as they typically require more CD and NN calls than non-optimal planners. We study the ratio between the overall time spent on NN search and CD after $N$ configurations were sampled. We observe situations where NN takes up to 100% more time than CD in scenarios based on the Open Motion Planning Library [15]; on synthetic high-dimensional C-spaces we even observe a ratio of 4500%.

We mostly concentrate on the *radial* version of MP algorithms, where the set of neighbors in the roadmap of a given configuration $q$ includes all configurations of maximal distance $r$ from $q$. To do so in *non-Euclidean* C-spaces, we derive closed-form expressions for the volume of a unit ball in several common C-spaces. This technical result is of independent interest, as the lack of such expressions seems to have thus far prevented the exploration and understand-ing of these types of algorithms in non-Euclidean settings—most experimental evaluation reported in the literature on the radial version of asymptotically-optimal planners is limited to Euclidean settings only. We show empirically that in certain scenarios, the radial version of an MP algorithm produces a solution of specific cost more than *ten times faster* than the non-radial version, namely, where each node is connected to its $k$ nearest neighbors.

We emphasize that we are not the first to claim that in certain cases NN may dominate the running of MP algorithms, see, e.g., [9]. However, we take a systematic approach to characterize and analyze when this phenomenon occurs.

Throughout the paper we use the following notation: For an algorithm ALG, let $\chi_{\text{ALG}}(S)$ be the ratio between the overall time spent on NN search and CD for a specific motion-planning problem after a set $S$ of configurations was sampled, where we assume that all other parameters of the problem, the workspace and the robot, are fixed—see details below. Let $\chi_{\text{ALG}}(N)$ be the expected value of $\chi_{\text{ALG}}(S)$ over all sample sets $S$ of size $N$.

*Organization* We start with an overview of related work in Sec. 2 and continue in Sec. 3 to summarize, for several algorithms, the computational complexity in terms of NN search and CD. We show that asymptotically, as $N$ tends to infinity, $\chi_{\text{ALG}}(N)$ tends to infinity as well. In Sec. 4 we point out several NN-sensitive settings together with simulations demonstrating how $\chi_{\text{ALG}}(N)$ behaves in such settings. These simulations make use of the closed-form expressions of the volume of unit balls, which are detailed in Sec. 5. Sec. 6 concludes with a discussion and possible future work.

**Remark.** There are algorithms where a third type of operations (e.g., cost estimation), beyond CD and NN, takes the lion's share of computation time. We defer the discussion of such algorithms to future research.

## 2 Background and related work

We start by giving an overview of asymptotically (near) optimal MP algorithms and continue with a description of CD and NN algorithms.

### 2.1 Asymptotically optimal sampling-based motion planning

A random geometric graph (RGG) $\mathcal{G}$ is a graph whose vertices are sampled at random from some space $\mathcal{X}$. Every two configurations are connected if their distance is less than a connection radius $r_n$ (which is typically a function of the number of nodes $n$ in the graph). We are interested in a connection radius such that, asymptotically, for any two vertices $x, y$, the cost of a path in the graph connecting $x$ and $y$ converges to the minimal-cost path connecting them in $\mathcal{X}$. A sufficient condition to ensure this property is that [27]

$$r_n \geq 2\eta \left(\frac{\mu(\mathcal{X}_{\text{free}})}{\zeta_d}\right)^{1/d} \left(\frac{1}{d}\right)^{1/d} \left(\frac{\log n}{n}\right)^{1/d}. \tag{1}$$

Here $d$ is the dimension of $\mathcal{X}$, $\mu(\cdot)$ and $\zeta_d$ denote the Lebesgue measure (volume) of a set and of the $d$-dimensional unit ball, respectively, and $\eta \geq 1$ is a tuning parameter that allows to balance between exploring unvisited regions of the C-space and connecting visited regions. Alternatively, an RGG where every vertex is connected to its $k_n \geq e(1 + 1/d) \log n$ nearest neighbors will ensure similar convergence properties [28]. Unless stated otherwise, we focus on RGGs of the former type. Namely, where the set of neighbors of a node is chosen according to Eq. 1. For a survey on additional models of RGGs, their properties and their connection to sampling-based MP algorithms, see [50].

Most asymptotically-optimal planners sample a set of collision-free configurations (either incrementally or in batches). This set of configurations induces an RGG $\mathcal{G}$ or a sequence of increasingly dense RGGs $\{\mathcal{G}_n\}$ whose vertices are the sampled configurations. Set $\mathcal{G}' \subseteq \mathcal{G}$ to be the subgraph of $\mathcal{G}$ whose edges represent collision-free motions. These algorithms construct a roadmap $\mathcal{H} \subseteq \mathcal{G}'$.

PRM* and RRG [28], which were two of the three initial asymptotically-optimal MP algorithms, call the local planner for *all* the edges of $\mathcal{G}$. To increase the convergence rate to high-quality solutions, algorithms such as RRT* [28], RRT# [3], LBT-RRT [47], FMT* [27], MPLB [46], Lazy-PRM* [20], and BIT* [18] call the local planner for a *subset* of the edges of $\mathcal{G}$.

Reducing the number of LP calls is typically done by constructing $\mathcal{G}$ (using nearest-neighbor operations only) and deciding for which edges to call the local planner. Many of the algorithms mentioned do so by using graph operations such as shortest-path computation. These operations often take a tiny fraction of the time required for LP computation. One such example is MPLB which uses shortest-path computation to obtain a lower bound on the cost-to-go of nodes. Alternatively, dynamic shortest-path algorithms such as LPA* [34] are used to maintain the cost-to-come of nodes when the graph undergoes a series of edge insertions and deletions. Examples include Lazy-PRM* and LBT-RRT. In more recent algorithms, such as FMT* and BIT*, NN and CD may not take the lion's share of computation time in practice.

Variants and extensions of the above algorithms include, among others, constructing sparse data structures [16, 48], taking into account dynamic obstacles [43] and addressing kinodynamic planning [38, 49, 53].

## 2.2 Collision detection

Collision-detection algorithms are extensively used by sampling-based MP algorithms in order to answer discrete collision queries or continuous ones. The former test whether a sampled configuration is in collision with the workspace obstacles, whereas the latter test whether a continuous path between two configurations is collision-free.

Most CD algorithms are bound to certain types of models, where rigid polyhedral models are the most common. They often allow answering proximity queries as well (i.e., separation-distance computation or penetration-depth estimation). Several software libraries for collision detection are publicly available [14, 35]. The most general of which is the Flexible Collision Library (FCL) [44] that integrates several techniques for fast and accurate collision checking and proximity computation. For polyhedral models, which are prevalent in MP settings, most commonly-used techniques are based on *bounding volume hierarchies* (BVH). Among the various types of bounding volumes, the most prominent are axis-aligned bounding box, oriented bounding box, spheres, and swept sphere volume. The different types of bounding volumes differ in the tightness of fitting to the bounded set and in the cost of an overlap test of two bounding volumes.

A collision query using BVHs may take $O(m^2)$ time in the worst case, where $m$ is the complexity of the obstacle polyhedra (recall that we assume that the robot system has constant-description complexity). However, tighter bounds may be obtained using methods tailored for large environments [14, 22]. Specifically, the time complexity is $O(m \log^{\delta-1} m + s)$, where $\delta \in \{2, 3\}$ is the dimension of the workspace $\mathcal{W}$ and $s$ is the number of intersections between the bounding volumes. Other methods relevant to MP are continuous CD [29, 30, 55] and algorithms tailored for dynamic environments where the objects undergo rigid motion [14]. Additionally, efficient approaches that use graphics hardware may accelerate the running time of CD queries significantly; see, e.g., [19]. For a survey on the topic, see [39].

### 2.3 Nearest-neighbor methods: exact and approximate

Nearest-neighbor (NN) algorithms are frequently used in various domains. In the most basic form of the problem we are given a set $P$ of $n$ points in a metric space $M = (X, \rho)$, where $X$ is a set and $\rho : X \times X \to \mathbb{R}$ is a distance metric. Given a query point $q \in X$, we wish to efficiently report the nearest point $p \in P$ to $q$. Immediate extensions include the $k$-nearest-neighbors (K-NN) and the $r$-near-neighbors (R-NN) problems. The former reports the $k$ nearest points of $P$ to $q$, whereas the latter reports all points of $P$ within a distance $r$ from $q$. Another variant is the all-pairs $r$-near neighbors (AP) where, given a radius $r$, one has to report all pairs of points in $P$ of distance at most $r$.

In the plane, the NN search problem (also known as the *post-office problem*) can be efficiently solved by constructing a Voronoi diagram of $P$ in $O(n \log n)$ time and preprocessing it to a linear-size point-location data structure in $O(n \log n)$ time[3]. Queries are then answered in $O(\log n)$ time [7, 21]. However, for high-dimensional point sets this approach becomes infeasible, as it is exponential in the dimension $d$. This phenomenon is often termed "the curse of dimensionality" [26].

An efficient data structure for low dimensional spaces[4] is the $kd$-tree [6, 17], whose expected query complexity is logarithmic in $n$ under certain assumptions. However, the constant factors hidden in the asymptotic query time depend exponentially on the dimension $d$ [5]. Another structure suitable for low-dimensional spaces is the geometric near-neighbor access tree (GNAT); as claimed in [11], typically the construction time is $O(dn \log n)$ and only linear space is required.

---

[3] A simple randomized algorithm with expected $O(n \log n)$ time is described in [21]. There is a variety of more involved algorithms with worst-case $O(n \log n)$ time (see, e.g., [31]).
[4] We refer to a space as low dimensional when its dimension is at most a few dozens.

In order to overcome the so-called "curse of dimensionality", numerous algorithms, such as, cover trees [8] and randomly-oriented $kd$-trees [51], were proposed. These methods adapt to the intrinsic dimension of the data, which is often much smaller than that of the ambient dimension $d$.

All the aforementioned structures give an exact solution to the problem. However, many approximate algorithms exist, and often perform significantly faster than the exact ones, especially when $d$ is high. The approximate variant of the basic NN search problem is to design a data structure that supports the following operation: For any query $q \in X$, if there exists $p \in P$ such that $D(p, q) \leq r$, find a point $p' \in P$ such that $D(q, p') \leq cr$, where $r, c > 0$ are given real parameters. This problem is often referred to as $(r, c)$-NN [25]. The approximate $k$-nearest neighbors and the approximate $r$-near neighbors search problems can be defined similarly. Among the prominent approximate algorithms are *Balanaced box-decomposition trees* (BBD-trees) [5], and Locality-sensitive hashing (LSH) [26]. An improved variant of LSH, which uses data-dependent hashing, was recently presented [2]. For a survey on approximate NN methods in high-dimensional spaces, see [25].

In the context of MP, several specifically-tailored exact [24, 54] and approximate [32, 45] techniques were previously described. Note that the proofs provided for the probabilistic completeness and asymptotic optimality of certain planners (e.g., PRM*, RRT*) assume the use of exact NN queries [28]. Conveniently, in [50] a theoretical justification for using approximate NN methods rather than exact ones is proven for PRM*.

A structure based on random grids allows for efficiently answering approximate $r$-near neighbors queries in Euclidean $d$-dimensional space [1]. In [32] we used this structure to obtain significant speedups in the construction time of certain asymptotically-optimal algorithms. In addition, we were able to converge faster to the optimal solution, when the grid-based structure was used to answer NN queries.

## 3 The asymptotic behavior of common MP algorithms

In this section we provide more background on the asymptotic complexity analysis of various sampling-based MP algorithms. We then show that for both PRM-type algorithms and RRT-type algorithms, the expected ratio between the time spent on NN search and the time spent on CD goes to infinity as $n \to \infty$.

Throughout the paper we use the following notation: We denote by $N$ the total number of configurations sampled by the algorithm, and by $n$ the number of collision-free configurations in the roadmap. Let $m$ denote the complexity of the workspace obstacles and assume that the robot is of constant-description complexity[5].

| Operation | Computational complexity | Comments |
|---|---|---|
| NN | $O(c_{d,\varepsilon} \cdot \log n)$ | Approx. using BBD-trees [4] |
| | $\Omega(\log n)$ | Approx. under the partition trees paradigm* |
| R-NN | $O(c'_{d,\varepsilon} + 2^d \log n + \kappa)$ | Approx. using BBD-trees [4] |
| | $\Omega(\log n + \frac{1}{\epsilon^{d-1}} + \kappa)$ | Approx. under the partition trees paradigm* |
| AP | $O(c_{\text{AP}} \cdot \log n \cdot (n + \kappa))$ | Approx. using RTG [1] |
| CD (for a robot with a single rigid part) | $O(m \log^{\delta-1} m + s)$ | Assuming the workspace is $\mathbb{R}^\delta$ [39] |
| CD (for a system with $\ell$ rigid parts)** | $O(\ell^2 + \ell m \log^{\delta-1} m + s)$ | Assuming the workspace is $\mathbb{R}^\delta$ [39] |

**Table 1:** Summary of the complexity of typical primitive operations in sampling-based algorithms. Here, $\kappa$ denotes the expected number of neighbors returned by an NN query, $\delta$ denotes the dimension of the workspace, and $s$ denotes the number of intersections between the bounding volumes. The constants $c_{d,\varepsilon}$, $c'_{d,\varepsilon}$ and $c_{\text{AP}}$ are (at least) exponential in $d$. RTG is an approximate NN structure based on random grids.
* The lower bounds for NN and R-NN are for the worst case query based on the partition trees paradigm.
** CD for a system with $\ell$ rigid parts involves $\ell$ collision checks between parts and obstacles and $O(\ell^2)$ collision checks between the pairs of parts.

### 3.1 Complexity of common motion-planning algorithms

We start by summarizing the computational complexity of the primitive operations and continue to detail the computational complexity of a selected set of algorithms. We assume familiarity with the planners that are discussed.

**Complexity of primitive operations** The main primitive operations that we consider are (i) nearest-neigbhor operations (NN, R-NN and AP) and (ii) collision-detection operations (CD and LP). Additionally, MP algorithms make use of priority queues and graph operations. We assume, as is typically the case, that the running time of these operations is negligible when compared to NN and CD.

Since many NN data structures require a preprocessing phase, the complexity of a single query should consider the amortized cost of preprocessing. However, since usually at least $n$ NN or R-NN queries are performed, where $n$ is the number of points stored in the NN data structure, this amortized preprocessing cost is asymptotically subsumed by the cost of a query.

---

[5] The assumption that the robot is of constant-description complexity implies that testing for *self-collision* can be done in constant time.

Table 1 summarizes the complexity of the main primitive operations. Note that the bounds given for NN assume approximate methods[6]. Local planning (LP), which is not mentioned in Table 1, is often implemented using multiple CD operations along a densely-sampled C-space line-segment between two configurations. Specifically, we assume that the planner is endowed with a fixed parameter called STEP specifying the sampling density along edges. During LP, edges of maximal length $r_n$ will be subdivided into $\lceil r_n/\text{STEP} \rceil$ collision-checked configurations (see also [36, p. 214]). Therefore, the complexity of a single LP query can be bounded by $O(r_n \cdot Q_{\text{CD}})$, where $Q_{\text{CD}}$ is the complexity of a single CD query (here STEP is assumed to be constant).

**Remark.** Our analysis differs slightly from the one performed by Karaman and Frazzoli [28] where LP is performed by checking if the C-space line segment intersects the C-space obstacles. The latter requires an exact representation of the C-space obstacles, whose complexity can be exponential in the dimension $d$. We therefore assume, as is common in practice, that LP is performed as a series of CD calls in the workspace.

**Complexity of algorithms** In order to choose which edges of $\mathcal{G}$ to explicitly check for being free using the local planner, all algorithms need to determine (i) which of the $N$ nodes are collision free and (ii) what are the neighbors of each node. Thus, these algorithms typically require $N$ CD calls[7] and either $n$ R-NN calls in the case of incremental algorithms, such as RRT*, LBT-RRT or RRT#, or one AP call[8] in the case of batch algorithms, such as sPRM*, Lazy-sPRM* or FMT*.

To quantify the number of LP calls performed by each algorithm, note that the expected number of neighbors of a node in $\mathcal{G}$ is $\Theta(\eta^d 2^d \log n)$ [50]. Therefore, if an algorithm calls the local planner for all (or for a constant fraction of) the edges of $\mathcal{G}$, then the expected number of LP calls will be $\Theta(\eta^d 2^d n \log n)$. Often, tighter bounds on the number of LP calls can be obtained. For instance, for FMT* (and similarly for MPLB) the expected number of LP calls can be bounded by $O(n)$ (see, [27, Lemma C.2]).

### 3.2 The asymptotic behavior of the ratio $\chi_{\text{ALG}}(N)$

Let $T_{\text{CD}}(S)$ be the overall time spent on CD for a specific motion-planning problem after a set $S$ of configurations was sampled, where we assume, as before, that all other parameters of the problem are fixed. Let $T_{\text{CD}}(N)$ be the expected value of $T_{\text{CD}}(S)$ over all sample sets $S$ of size $N$. We show here that the expected value of the ratio $\chi_{\text{ALG}}(N)$ over all sample sets of size $N$ goes to infinty as $N \to \infty$ for both sPRM* and RRT*. Recall that we are interested in the expected value of the ratio. We do that by looking at the ratio between a lower bound on the time of NN and $T_{\text{CD}}(N)$, defined above.

---

[6] Bounds for exact NN structures exist only for a subset of the prevalent methods and may require prior assumptions on the point set.

[7] To determine if a node is collision-free, one may use the framework by Bialkowski et al. [9] which replaces CD calls with NN search. However, their framework assumes that the CD provides a bound on the distance to the closest C-space obstacle, an assumption that may not always hold. Thus, here we assume standard CD implementation.

[8] AP can be replaced by a series of $n$ R-NN queries.

To obtain a lower bound on the time of NN, we assume that the NN structure being used is a $j$-ary tree for a constant $j$, in which the data points are kept in the leaves. This is a reasonable assumption, as many standard NN structures are based on trees [5, 6, 11, 17]. Performing $n$ queries of NN (or R-NN) using this structure, one for every data point, costs $\Omega(n \log n)$, as each query involves locating the leaf in which the query point lies. It is easy to show this both for sPRM*, in which the NN structure is constructed given a batch of all data points, and for RRT*, where the structure is constructed incrementally.

Additionally, we have the following lemma, whose proof is in [33]:

**Lemma 1** *If an algorithm uses a uniform set of samples and the C-space obstacles occupy a constant fraction of the C-space, then $n = \Theta(N)$ almost surely.*

*Proof.* Lemma 1 states that under mild assumptions, there exist constants $N_0 \in \mathbb{N}^+$ and $0 < c_1 \leq 1$ s.t. for every $N > N_0$, $n \geq c_1 N$ almost surely. Showing that this holds for sPRM* requires using the Chernoff bound (see, e.g., [42, Theorem 4.5]) and noting that the number $n$ is a binomial random variable $B(N, p)$ with success probability $p = \mu(\mathcal{X}_{\text{free}})/\mu(\mathcal{X})$, which is a strictly-positive constant by our assumption. In particular, for every constant $\delta \in (0, 1)$ the event $n > (1 - \delta)pN$ holds almost surely. In order to show that $n$ is proportional to $N$ in RRT* as well, we supplement the latter observation with the fact that after a finite number of samples, each new free sample will be added as a node to the constructed tree. This is formalized in [28, Lemma 63]. □

We use the following notation in the remainder of this section: For an operation OP let $Q_{\text{OP}}$ denote the complexity of the operation and $\#_{\text{OP}}$ denote the number of times the operation is called.

**The asymptotic value of $\chi_{\text{sPRM*}}(N)$** For sPRM* it holds that $T_{\text{CD}}(N) = \#_{\text{CD}} \cdot Q_{\text{CD}} + \#_{\text{LP}} \cdot Q_{\text{LP}}$. Clearly, $\#_{\text{CD}} = N$ and $Q_{\text{CD}} = O(m^2)$ (an upper bound on $s$ from Table 1—the number of intersections between the bounding volumes). In expectation we have that $\#_{\text{LP}} = O(r_n^d)$ and in addition $Q_{\text{LP}} = O(r_n \cdot Q_{\text{CD}})$. Finally, recall that $r_n = \Theta\left(2\eta (\log n/n)^{1/d}\right) = \Theta\left((\log n/n)^{1/d}\right)$. Therefore,

$$\begin{aligned} T_{\text{CD}}(N) &= N \cdot Q_{\text{CD}} + O(\eta^d 2^d n \log n) \cdot Q_{\text{LP}} \\ &= N \cdot Q_{\text{CD}} + O(\eta^d 2^d n \log n) \cdot O(r_n \cdot Q_{\text{CD}}) \\ &= O(m^2 N) + O(m^2 N^{1-1/d} \log^{1+1/d} N) = O(m^2 N). \end{aligned} \quad (2)$$

As $\Omega(n \log n)$ is a valid lower bound on the overall complexity of NN, there exists a constant $c_2 > 0$ s.t. the time for NN for a roadmap with $n$ nodes is at least $c_2 n \log n$. Moreover, since $T_{\text{CD}}(N) = O(m^2 N)$ then there exists a constant $c_3 > 0$ s.t. the overall time for CD is at most $c_3 m^2 N$.

Thus, using Lemma 1,

$$\chi_{\text{sPRM}^*}(N) \geq \frac{c_2 n \log n}{c_3 m^2 N} \geq \frac{c' \log N}{m^2}, \qquad (3)$$

where $c' > 0$ is a constant. Observing that the above fraction goes to infinity as $N$ goes to infinity, we obtain that $\lim_{N \to \infty} \chi_{\text{sPRM}^*}(N) = \infty$, as anticipated.

We note that although $m$ is assumed to be constant, we leave it in our analysis to emphasize its effect on $\chi_{\text{ALG}}(N)$.

**The asymptotic value of $\chi_{\text{RRT}^*}(N)$** Since RRT* is an incremental algorithm and not a batch one, we consider the time spent on CD in the $N$th iteration, which is $\text{Q}_{\text{CD}} + \#_{\text{LP}} \cdot \text{Q}_{\text{LP}}$. Therefore,

$$\text{T}_{\text{CD}}(N) = \sum_{i=1}^{N} \left( O(m^2) + O(\eta^d 2^d \log i (\log i/i)^{1/d}) \cdot O(m^2) \right)$$
$$= O(m^2 N) + O\left( m^2 \eta^d 2^d \sum_{i=1}^{N} \left( \log i \left( \frac{\log i}{i} \right)^{1/d} \right) \right).$$

Note that since $f(i) = \log i (\log i / i)^{1/d}$ is bounded from above by some constant $b$, then $\sum_{i=1}^{N} f(i)$ can be upper bounded by $O(Nb)$. Thus, $\text{T}_{\text{CD}}(N) = O(m^2 N)$ and, accordingly, $\chi_{\text{RRT}^*}(N) \geq \frac{c'' \log N}{m^2}$, where $c'' > 0$ is a constant. Therefore, $\lim_{N \to \infty} \chi_{\text{RRT}^*}(N) = \infty$. In summary,

**Proposition 2** *The values $\chi_{sPRM^*}(N)$ and $\chi_{RRT^*}(N)$ tend to infinity as $N \to \infty$.*

From a theoretical standpoint, NN search determines the asymptotic running time of typical sampling-based MP algorithms. In contrast, the common experience is that CD dominates the running time in practice. However, we show in the remainder of the paper that in a variety of special situations NN search is a non-negligible factor in the running-time in practice.

## 4 Nearest-neighbor sensitive settings

In this section we describe settings where the computational role of NN search in practice is far from negligible, even for a relatively small number of samples. We call these settings *NN-sensitive*. For each such setting we empirically demonstrate this behavior. In Sec. 4.1 we describe our experimental methodology and outline properties common to all our experiments. Each of the subsequent sections is devoted to a specific type of NN-sensitivity.

### 4.1 Experimental methodology

In our experiments, we ran the Open Motion Planning Library (OMPL 1.1) [15] on a 2.5GHz×4 Intel Core i5 processor with 8GB of memory. Each reported result is averaged over fifty (50) runs and includes error bars which denote the 20'th and 80'th percentiles. The scenarios used are depicted in Fig. 2.

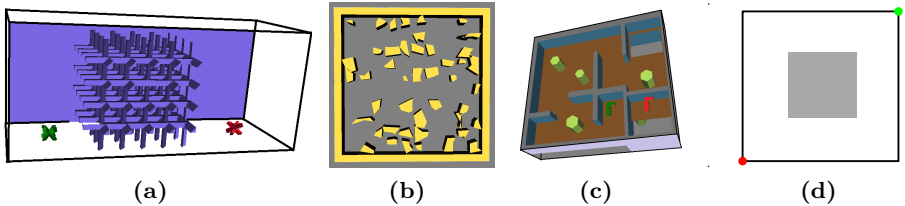

**Fig. 2:** Scenarios used in experiments. (a) 3D Grid, (b) 2D Random polygons, (c) 3D Cubicles and (d) $d$D Hypercube with a centered obstacle. Start and target configurations for a robot are depicted in green and red, respectively. Scenarios (b) and (c) are provided with the OMPL distribution. More details are provided in the body of the paper.

Several of our experiments are in non-Euclidean C-spaces, which in turn require a closed-form expression for $\zeta_d$, the measure (volume) of the $d$-dimensional unit ball (see Eq. 1). In Sec. 5 we describe a general approach to compute this value together with a heuristic that makes the computed radius effective in practice. This heuristic is used in all the experiments presented in this section.

*What do we measure?* Recall that our main thesis in this paper is that while the folklore in motion planning is that the running time of sampling-based planning algorithms in practice is strongly dominated by collision detection, we (and others) observe that quite often the time taken up by NN-search is significant, and not rarely larger than the time taken up by collision detection. Therefore, our primary measure is *wall time*, namely the running time spent on the different primitives as gauged in standard clock time (to distinguish from CPU-time measurement or other more system-specific measurements like number of floating point operations). The principal reason for doing that is that wall time is what matters most in practice. This, for example, will affect the response time of a planner used by a robot system. One may argue that this measurement may only be meaningful for a very limited suite of software tools used by motion planners. However, we use state-of-the-art tools that are used by many. There is not such an abundance of efficient stable software tools for this purpose, and most researchers in the field seem to use a fairly small set of tools for collision detection and nearest-neighbor search. This said, we still provide additional measurements for each experiment—the average number of basic operations. These measurements should allow people who come up with their own (or specialized) motion-planning primitives to assess what will be the effect of their special primitives on the overall running time of the algorithms in practice.

### 4.2 NN-sensitive algorithms

In recent years, several planners were introduced, which *algorithmically* shift some of the computational cost from CD to NN search. Two such examples are Lazy-PRM* [20] and MPLB [46], though lazy planners were described before (e.g., [10]). Both algorithms delay local planning by building an RGG $\mathcal{G}$ over a set of samples *without* checking if the edges are collision free. Then, they employ graph-search algorithms to find a solution. To construct $\mathcal{G}$ only NN queries are required. Moreover, using these graph-search algorithms dramatically reduces the number of LP calls. Thus, in many cases (especially as the number of samples grows) the weight of CD is almost negligible with respect to that of NN.

Specifically, Lazy-PRM* iteratively computes the shortest path in $\mathcal{G}$ between the start and target configurations using a dynamic single source shortest path algorithm. LP is called only for the edges of the path. If some are found to be in collision, they are removed from the graph. This process is repeated until a solution is found or until the source and target do not lie in the same connected component. We use a batch variant of Hauser's Lazy-PRM* algorithm [20], which we denote by Lazy-sPRM*. This variant constructs the roadmap in the same fashion as sPRM* does but delays LP to the query phase.

MPLB uses $\mathcal{G}$ to compute lower bounds on the cost between configurations to tightly estimate the cost-to-go [46]. These bounds are then used as a heuristic to guide the search of an anytime version of FMT* [27]. The bounds are computed by running a shortest-path algorithm over $\mathcal{G}$ from the target to the source. Fig. 1 (on page 2) presents the amount of NN, CD and other operations used by MPLB running on the 3D Grid scenario for two robots translating and rotating in space that need to exchange their positions (Fig. 2a). Fig. 3 plots the ratio $\chi_{\text{MPLB}}(N)$ as a function of the number of valid samples $n$ and Table 2 provides the average number of basic operation calls performed for different values of $n$. Clearly, with several thousands of iterations, which are required for obtaining a high-quality solution, NN dominates the running time of the algorithm.

Additional experiments demonstrating the behavior of NN-sensitive algorithms can be found in [9, 20, 46].

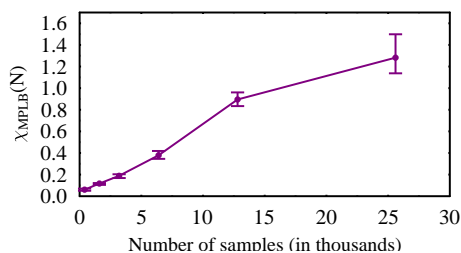

| $n$ | $\#_{\text{R-NN}}$ | $\#_{\text{CD}}$ | $\#_{\text{LP}}$ | $\#_{\text{CD in LP}}$ |
|---|---|---|---|---|
| 1,600 | 3.1K | 4.2K | 9.5K | 159K |
| 3,200 | 6.3K | 8.5K | 20.8K | 308K |
| 6,400 | 12.7K | 16.9K | 44.2K | 598K |
| 12,800 | 25.5K | 33.9K | 86.7K | 1,105K |

**Fig. 3:** $\chi_{\text{MPLB}}(\mathsf{N})$ as a function of $n$ demonstrating an NN-sensitive algorithm.

**Table 2:** Average number of calls for the main primitive operations for different values of $n$ for MPLB.

### 4.3 NN-sensitive scenarios

A scenario $\mathcal{S} = (\mathcal{W}, \mathcal{R})$ is defined by a workspace $\mathcal{W}$ and a robot system $\mathcal{R}$. The robot system $\mathcal{R}$ may, in turn, be a set of $\ell$ single constant-description complexity robots operating simultaneously in $\mathcal{W}$. Let the dimension $d$ of $\mathcal{S}$ be the dimension of the C-space induced by $\mathcal{R}$, and, hence, $d = \Theta(\ell)$.[9] Let the complexity of $\mathcal{S}$ be the complexity $m$ of the workspace obstacles. Note that CD is affected by $\ell$, as both robot-obstacle and robot-robot collisions should be considered. Therefore, the bound on the complexity of a CD operation is: $O(\ell \cdot m^2 + \ell^2)$, see Sec. 3.1.

We next show how the role of NN may increase when (i) the dimension of $\mathcal{S}$ increases or (ii) the complexity of $\mathcal{S}$ decreases.

**The effect of the dimension $d$** In Section 3.2 we show that as the number of samples tends to infinity, NN dominates the running time of the algorithm. A natural question to ask is "what happens when we fix the number of samples and increase the dimension?" The different structure of RRT* and sPRM* merits a different answer for each algorithm.

*RRT\** Here, we show that the NN sensitivity grows with the number of *unsuccessful iterations*[10]. This implies that if the number of unsuccessful iterations grows with the dimension, so will $\chi_{\text{RRT}^*}(N)$. Indeed, we demonstrate this phenomenon in the 3D cubicles scenario (Fig. 2c). Note that in this situation the effect of $d$ is indirect.

To better discuss our results we define two types of LP operations: the first is called when the algorithm attempts to grow the tree towards a random sample while the second is called during the rewiring step. We denote the former type of LP calls by LP-A and the latter by LP-B and note that LP-A will occur every iteration while LP-B will occur only in successful ones.

We use $\ell$ translating and rotating L-shaped robots. We gradually increase $\ell$ from two to six, resulting in a C-space of dimension $d = 6\ell$. Robots are placed in different sections of the workspace and can reach their target with little robot-robot interaction. We fix the number of iterations $N$ and measure $\chi_{\text{RRT}^*}(N)$ as a function of $d$. The results for several values of $N$ are depicted in Fig. 4. Additionally, Table 3 shows the average number of operation calls for various values of $d$.

As $d$ grows, the number of unsuccessful iterations grows (see $\#_{\text{LP-A}}$ in Table 3). This growth, which is roughly linear with respect to $d$ induces a linear increase in $\chi_{\text{RRT}^*}(N)$ for a given $N$ (see Fig. 4). Furthermore, the slope of this line increases with $N$ which further demonstrates the fact that for a fixed $d$, $\lim_{N \to \infty} \chi_{\text{RRT}^*}(N) = \infty$.

Finally, Fig. 5 depicts the time decomposition of the main primitives as a function of $d$, for $N = 80K$. Clearly, as $d$ grows most of the time is spent on NN calls due to the increase in the portion of unsuccessful iterations.

---

[9] In the case of a single $\ell$-link robot, the robot is not of constant-description complexity. Thus, the dimension of $\mathcal{S}$ is $d = \Theta(\ell)$. For simplicity of exposition we ignore here the case of many non-constant multi-link robots.

[10] Here, an iteration is said to be unsuccessful when the RRT* tree is not extended.

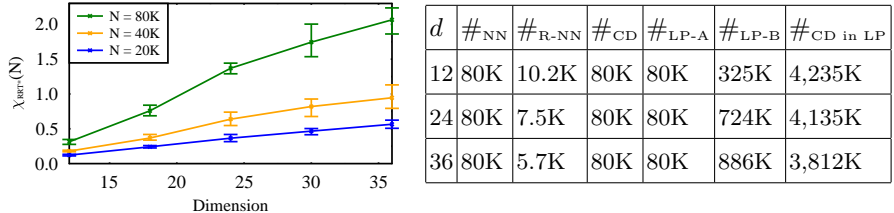

| $d$ | $\#_{\text{NN}}$ | $\#_{\text{R-NN}}$ | $\#_{\text{CD}}$ | $\#_{\text{LP-A}}$ | $\#_{\text{LP-B}}$ | $\#_{\text{CD in LP}}$ |
|---|---|---|---|---|---|---|
| 12 | 80K | 10.2K | 80K | 80K | 325K | 4,235K |
| 24 | 80K | 7.5K | 80K | 80K | 724K | 4,135K |
| 36 | 80K | 5.7K | 80K | 80K | 886K | 3,812K |

**Fig. 4:** $\chi_{\text{RRT}^*}(\mathsf{N})$ as a function of $d$ in the 3D Cubicles scenario (Fig. 2c), when fixing the number $N$ of iterations.

**Table 3:** Average number of calls for the main primitive operations for different values of $d$, for $N = 80K$ iterations.

*sPRM\** Here, the NN sensitivity of the algorithm is more complex. The reason, roughly speaking, is that for a *fixed* $n$, the expected value of the number $\kappa$ of reported neighbors is $\Theta(2^d n \log n)$. Thus, in expectation, $\kappa$ grows exponentially in $d$. However, for large enough values of $d$, we have $\kappa = \Theta(n^2)$, since each node can have at most $n$ neighbors. Interestingly, this means that the computational cost of the overall NN time shifts from *finding* the set of nearest neighbors to *reporting* it.

We attempt to formalize the above discussion but emphasize that this (crude) analysis is only useful to try and explain trends. The reason being is that we consider finite values of $n$ and $d$ while using asymptotic bounds. Clearly in this regime, constants may play a significant role, a fact that we neglect.

Set $\kappa$ to be the total number of nearest neighbors computed by the algorithm. Following the above discussion, for a fixed $n$ and increasing values of $d$, the expected value of $\kappa$ grows from $O(n \log n)$ to $O(n^2)$. As $\kappa$ grows, LP will dominate the running time of $T_{\text{CD}}(N)$ (and *not* CD as asymptotically occurs in 2). Furthermore, for large values of $d$, we have that $r_n \approx c$ for some constant $c$. Thus, $T_{\text{CD}}(N) \approx cn^2$. For NN, as we discussed, the computational cost will shift from *finding* the set of nearest neighbors to *reporting* it. Thus, for large values of $d$, the cost of NN will be proportional to $n^2$. Combining the two observations, we can conjecture that for large values of $d$, we have that $\chi_{\text{sPRM}^*}(N)$ will converge to some constant.

According to the above analysis we expect to see an initial increase of $\chi_{\text{sPRM}^*}(N)$ followed by a convergence to a constant value. The increase in $\chi_{\text{sPRM}^*}(N)$ is due to the increasing complexity of finding the set of nearest neighbors, which grows with the dimension[11]. A possible decrease will occur as the computational weight "shifts" to reporting the set of neighbors followed by an asymptotic convergence to a constant value, for large values of $d$.

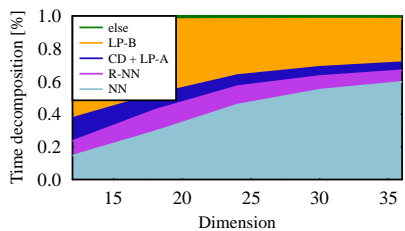

**Fig. 5:** Time breakdown of the main primitive operations in RRT* running on the 3D Cubicles scenario (Fig. 2c) as a function of $d$, for $N = 80$K iterations. Best viewed in color.

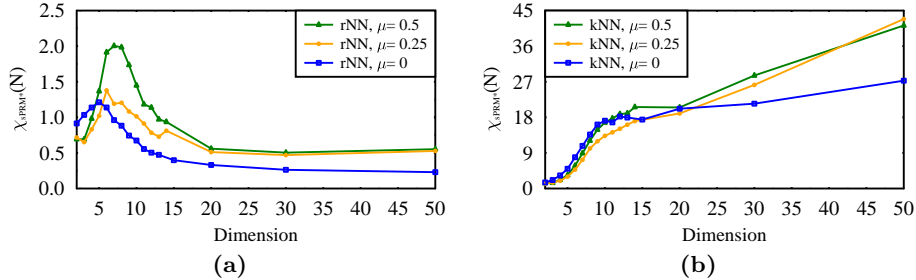

**Fig. 6:** $\chi_{\text{sPRM}^*}(N)$ as a function of $d$ in the $d$D Hypercube scenario (Fig. 2d) for a roadmap with $n = 5000$ vertices. Experiments with obstacle measures $\mu$ of $0, 0.25$ and $0.5$ are displayed. The connection strategies used in (a) and (b) are R-NN and K-NN, respectively.

Aiming to test this conjecture, we solved a planning problem for a point robot moving in the $d$-dimensional unit hypercube containing a hyper-cubicle obstacle of a specific measure $\mu$ centered at the middle of the workspace (Fig. 2d). Note that a single CD call in this setting takes $O(d)$ time. We set $n = 5000$ and used sPRM* to plan a path for a point robot from $(0, \ldots, 0) \in \mathbb{R}^d$ to $(1, \ldots, 1) \in \mathbb{R}^d$. We gradually increased $d$ and measured $\chi_{\text{sPRM}^*}(N)$. We repeated the experiment for obstacle volumes $\mu$: $0, 0.25$, and $0.5$, keeping the volume of the room fixed.

Indeed, the expected trend can be seen clearly in our results in Fig. 6a. We note, however, that the average number of neighbors of a node reached the value of 90%, only for $d \geq 30$. Moreover, for larger value of $\mu$ we saw a decrease in the average number of reported neighbors. This may explain the fact that the peak of the plot for $\mu = 0.5$ is obtained for a larger value of $d$, then the one for $\mu = 0$. Additionally, as we increase $\mu$, the ratio obtains higher values, before it starts converging. We repeated the experiment while using K-NN instead of R-NN queries. The results, depicted in Fig. 6b, show a growth in the ratio as a function of $d$ by 2000%. This is not surprising, as the standard value of $k$ that is commonly used is proportional to $\log n$, and is smaller by a factor of $2^d$ than the expected number of neighbors returned by an R-NN query.

**The effect of the geometric complexity $m$ of the obstacles** Recall that a collision query may take $O(m^2)$ in the worst case. For small values of $m$, this becomes negligible with respect to other primitive operations, such as NN. In order to demonstrate this effect we ran the following experiment which is based on the 2D Random polygons scenario (Fig. 2b). We created two sequences of increasing geometric-complexity (growing $m$) environments. Each sequence was constructed as follows: we start with the empty environment and incrementally add random polygons from Fig. 2b until all the polygons in Fig. 2b have been added. We then placed *eight* robots[12] that need to change their positions, and ran LBT-RRT (with approximation factor $\varepsilon = 0.4$) for a fixed $N$. Fig. 7 plots $\chi_{\text{LBT-RRT}}(N)$ as a function of $m$ for two sets of environments. As anticipated, the ratio in both sets of environments decays polynomially as $m$ grows. Average operation counts, given in Table 4, reveal an increase in the number of LP calls as $m$ grows. A possible explanation is that when there are very few polygonal obstacles in this environment, almost no rewiring is performed as the cost-to-come of each node in the LBT-RRT tree is close to optimal. However, the number of CD calls within LP operations decreases as $m$ grows since many edges are invalid. Note that each CD call becomes more costly as a function of $m$, thus, the overall time spent on CD does not change significantly as a function of $m$. Finally, since it is more difficult to grow the roadmap in the presence of obstacles, less R-NN calls are initiated, thus, reducing the time spent on NN.

---

[11] Here, we assume, as is common in the literature, that the cost of *finding* the set of NN grows with the dimension.

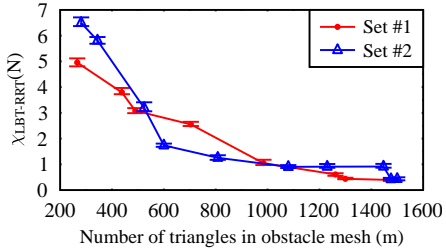

| $m$   | $\#_{\text{NN}}$ | $\#_{\text{R-NN}}$ | $\#_{\text{CD}}$ | $\#_{\text{LP}}$ | $\#_{\text{CD in LP}}$ |
|-------|------|--------|-----|-------|-------|
| 266   | 4K   | 2.5K   | 4K  | 8.1K  | 496K  |
| 704   | 4K   | 1.7K   | 4K  | 7.7K  | 352K  |
| 1,262 | 4K   | 0.99K  | 4K  | 24.5K | 271K  |
| 1,476 | 4K   | 0.7K   | 4K  | 28.7K | 232K  |

**Fig. 7:** $\chi_{\text{LBT-RRT}}(\mathsf{N})$ as a function of $m$ in workspaces of increasing obstacle complexity based on the 2D Random polygons scenario (Fig. 2b). The presented plots are for $N = 4\text{K}$ iterations in two different randomly-generated experiment sets.

**Table 4:** Average number of calls for the main primitive operations for different values of $m$ (chosen arbitrarily).

### 4.4 NN-sensitive parameters

Typically, when MP algorithms are evaluated, this is done after a careful tuning of their parameters. However, in practice, certain algorithms might be used with a sub-optimal set of parameters. This may be either due to lack of knowledge, errors, or the inability to choose the right set of parameters. In all planning algorithms, one of the critical user-defined parameters, is the step size (STEP); see Sec. 3. Using STEP which is too small may cause LP to be over-conservative and costly. Choosing larger values which are still appropriate for the scenario at hand allows to decrease the portion of time spent on CD checks.

---

[12] We use eight robots, as this is the smallest number of robots where the trend is easily discernible.

We demonstrate how RRT* becomes NN-sensitive under certain step-size values, by testing the effect of the step size on $\chi_{\text{RRT}^*}(N)$. We ran RRT* for a fixed number of iterations $N = 25K$ on the 3D Cubicles scenario (Fig. 2c). In order to modify the step size in OMPL, one needs to specify a state validity-checking resolution. This value, which we denote by RES, is specified as a fraction of the space's extent, that is, the maximal possible distance between two configurations in the C-space. Using larger values of RES may yield paths that are invalid. Thus, when increasing RES, we also used a model of the robot which was inflated accordingly to ensure that all paths are collision free (see [36, Ch.5.3.4]). One can see (Fig. 8) that there is a linear correlation between RES and $\chi_{\text{RRT}^*}(N)$. Indeed, Table 5 shows that when RES increases by a factor of 2, the number of CD calls within LP operations decreases by roughly 2. Maybe more interesting is that using the default OMPL value of 1%, CD takes roughly twenty times more than NN. By changing this value (and also using an inflated model of the robot), CD takes less than three times the amount of time spent on NN.

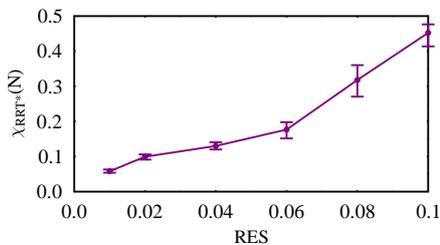

| RES | $\#_{\text{NN}}$ | $\#_{\text{R-NN}}$ | $\#_{\text{CD}}$ | $\#_{\text{LP}}$ | $\#_{\text{CD in LP}}$ |
|---|---|---|---|---|---|
| 0.01 | 25K | 2.9K | 25K | 36.1K | 715K |
| 0.02 | 25K | 3K | 25K | 36.5K | 384K |
| 0.04 | 25K | 3K | 25K | 36.4K | 197K |

**Fig. 8:** $\chi_{\text{RRT}^*}(N)$ as a function of the state validity-checking resolution RES for RRT* running on the 3D Cubicles scenario (Fig. 2c).

**Table 5:** Average number of calls for the main primitive operations for different values of $m$ (chosen arbitrarily).

## 5 Asymptotically-optimal motion-planning using R-NN

In this section we address an existing gap in the literature of sampling-based MP algorithms: How to use Eq. 1 in non-Euclidean spaces, which are prevalent in motion planning. Specifically, we derive closed-form expressions for the volume of the unit ball in several common C-spaces and distance metrics and discuss how to effectively use this value.

Closing this gap allows to evaluate the connection scheme of an algorithm. Namely, should one choose connections using R-NN or K-NN. In NN-sensitive settings this choice may have a dramatic effect on the performance of the algorithm since (i) the number of reported neighbors may differ and (ii) the cost of the two query types for a certain NN data structure may be different. Indeed, we show empirically that there are scenarios where using R-NN, a solution path of a specific cost is produced ten-fold (and more) faster than with K-NN.

### 5.1 Well-behaved spaces and the volume of balls

Recall that $\mathcal{X}$ denotes a C-space and that given a set $A \subseteq \mathcal{X}$, $\mu(A)$ denotes the Lebesgue measure of $A$. Let $\rho : \mathcal{X} \times \mathcal{X} \to \mathbb{R}$ denote a distance metric and let $\mathcal{B}^\rho_\mathcal{X}(r, x) := \{y \in \mathcal{X} | \rho(x, y) \leq r\}$ and $\mathcal{S}^\rho_\mathcal{X}(r, x) := \{y \in \mathcal{X} | \rho(x, y) = r\}$ denote the ball and sphere of radius $r$ (defined using $\rho$) centered at $x \in \mathcal{X}$, respectively. Finally, let $\mathbb{B}^\rho_\mathcal{X}(r) := \mu\left(\mathcal{B}^\rho_\mathcal{X}(r, 0)\right)$ and $\mathbb{S}^\rho_\mathcal{X}(r) := \mu\left(\mathcal{S}^\rho_\mathcal{X}(r, 0)\right)$. We will often omit the superscript $\rho$ or the subscript $\mathcal{X}$ when they will be clear from the context.

We now define the notion of a *well-behaved* space in the context of metrics; for a detailed discussion on well-behaved spaces see [40]. In such spaces there is a derivative relationship between $\mathbb{S}(r)$ and $\mathbb{B}(r)$. Formally,

**Definition 3** *A space $\mathcal{X}$ is well behaved when $\frac{\partial \mathbb{B}_\mathcal{X}(r)}{\partial r} = \mathbb{S}_\mathcal{X}(r)$. Conversely, we say that $\mathcal{X}$ is well behaved when $\int_{\varrho \in [0,r]} \mathbb{S}_\mathcal{X}(\varrho)d\varrho = \mathbb{B}_\mathcal{X}(r)$.*

We continue with the definition of a *compound space* which is the Cartesian product of two spaces. Let $\mathcal{X}_1, \mathcal{X}_2$ be two C-spaces with distance metrics $\rho_1, \rho_2$, respectively. Define $\mathcal{X} = \mathcal{X}_1 \times \mathcal{X}_2$ to be their compound space. We adopt a common way[13] to define the (weighted) distance metric over $\mathcal{X}$, when using weights $w_1, w_2 \in \mathbb{R}^+$ and some constant $p$ [36, Chapter 5]:

$$\rho_\mathcal{X} = \left(w_1 \rho_1^p + w_2 \rho_2^p\right)^{1/p}. \tag{4}$$

The following Lemma states that the volume of balls in a compound space $\mathcal{X} = \mathcal{X}_1 \times \mathcal{X}_2$ where $\mathcal{X}_1$ is well behaved can be expressed analytically.

**Lemma 4** *Following the above notation, if $\mathcal{X}_1$ is well behaved then*

$$\mathbb{B}_{\mathcal{X}_1 \times \mathcal{X}_2}(r) = \int_{\varrho \in [0, r/w_1^{1/p}]} \mathbb{S}_{\mathcal{X}_1}(\varrho) \cdot \mathbb{B}_{\mathcal{X}_2}\left(\left(\frac{r^p - w_1\varrho^p}{w_2}\right)^{1/p}\right) d\varrho. \tag{5}$$

*Proof.* By definition, $\mathbb{B}_\mathcal{X}(r) = \int_{x \in \mathcal{B}_\mathcal{X}(r)} dx$. Using Fubini's Theorem [41],

$$\mathbb{B}_{\mathcal{X}_1 \times \mathcal{X}_2}(r) = \int_{x_1 \in \mathcal{B}_{\mathcal{X}_1}(r/w_1^{1/p})} \left( \int_{x_2 \in \mathcal{B}_{\mathcal{X}_2}\left(\left(\frac{r^p - w_1 x_1^p}{w_2}\right)^{1/p}\right)} dx_2 \right) dx_1.$$

The inner integral is simply the volume of a ball of radius $\left(\frac{r^p - w_1 x_1^p}{w_2}\right)^{1/p}$ in $\mathcal{X}_2$. In addition, we know that $\mathcal{X}_1$ is well behaved, thus $\mathbb{B}_{\mathcal{X}_1}(r) = \int_{x_1 \in \mathcal{B}_{\mathcal{X}_1}(r)} dx = \int_{\varrho \in [0,r]} \mathbb{S}_{\mathcal{X}_1}(\varrho)d\varrho$. By changing the integration variable, substituting the inner integral and using the fact that $\mathcal{X}_1$ is well behaved we obtain Eq. 5. □

---

[13] Eq. 4 is often used due to its computational efficiency and simplicity. However, alternative methods exist, which exhibit favorable properties such as invariance to rotation of the reference frame; see, e.g., [12].

## 5.2 Volume of balls in common C-spaces

In this section we demonstrate how to use Lemma 4 for some common C-spaces. For a full list of C-spaces for which a closed-form expression for $\mathbb{B}_\mathcal{X}(r)$ was derived, we refer the reader to the appendix.

*SE(3)—The C-space of a spatial translating and rotating rigid body.* Recall that $\text{SE}(3) = \mathbb{R}^3 \times \text{SO}(3)$. We define the weighted distance metric of SE(3) according to Eq. 4 with $p = 1$ (a common choice in MP [15]). In the appendix we show that the following holds:

$$\mathbb{S}_{\mathbb{R}^3} = 4\pi r^2, \quad \mathbb{B}_{\text{SO}(3)}(r) = \pi(2r - \sin 2r)$$

Using the fact that $\mathbb{R}^3$ is well behaved and by Lemma 4 we obtain that:

$$\mathbb{B}_{SE(3)}(r) = \int_{\varrho \in [0, r/w_1]} (4\pi \varrho^2) \cdot \pi \left( \frac{2(r - w_1 \varrho)}{w_2} - \sin \frac{2(r - w_1 \varrho)}{w_2} \right) d\varrho$$

$$= \frac{\pi^2}{3} \frac{1}{w_1^3 w_2} \left( 2r^4 - 6w_2^2 r^2 + 3w_2^4 - 3w_2^4 \cos \frac{2r}{w_2} \right). \quad (6)$$

*SE(2)×SE(2)—The C-space of two translating and rotating planar robots.* The metric considered in this case is

$$\rho_{\text{SE}(2) \times \text{SE}(2)} = w_1 \rho_{\mathbb{R}^2} + w_2 \rho_{S^1} + w_3 \rho_{\mathbb{R}^2} + w_4 \rho_{S^1}.$$

We cannot compute the volume of a ball in a straightforward fashion as we do not have a representation of $\mathbb{S}_{\text{SE}(2)}(r)$. Thus, we start by computing the volume of a ball in $S^1 \times \text{SE}(2)$. In the appendix we show the following:

$$\mathbb{S}_{S^1}(r) = 2, \quad \mathbb{B}_{\text{SE}(2)}(r) = \frac{2\pi}{3} \frac{1}{w_3^2 w_4} r^3.$$

Since $S^1$ is well behaved, we obtain,

$$\mathbb{B}_{S^1 \times \text{SE}(2)}(r) = \int_{\varrho \in [0, r/w_2]} 2 \cdot \frac{2\pi}{3} \frac{1}{w_3^2 w_4} \left( \frac{r - w_2 \varrho}{1} \right)^3 d\varrho = \frac{\pi}{3} \frac{1}{w_2 w_3^2 w_4} r^4.$$

We now use the fact that $\text{SE}(2) \times \text{SE}(2) = \mathbb{R}^2 \times (S^1 \times \text{SE}(2))$ in order to compute the value of $\mathbb{B}_{\text{SE}(2) \times \text{SE}(2)}(r)$. As we show in the appendix, $\mathbb{S}_{\mathbb{R}^2}(r) = 2\pi r$ and $\mathbb{R}^2$ is well behaved. Thus, we obtain:

$$\mathbb{B}_{\text{SE}(2) \times \text{SE}(2)}(r) = \int_{\varrho \in [0, r/w_1]} 2\pi \varrho \cdot \frac{\pi}{3} \frac{(r - w_1 \varrho)^4}{w_2 w_3^2 w_4} d\varrho = \frac{\pi^2}{45} \frac{1}{w_1^2 w_2 w_3^2 w_4} r^6.$$

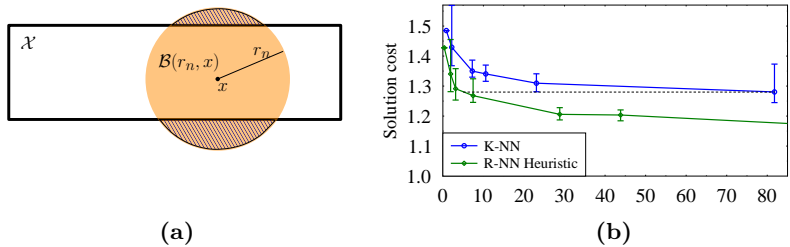

(a)    (b)

**Fig. 9:** R-NN heuristic (a) A two-dimensional C-space for which Assumption 5 does not hold. (b) Solution cost (optimal cost has a value of 1) as a function of time for Lazy-sPRM* running in the Cubicles scenario (Fig. 2c). The dashed line visualizes the difference in time for obtaining a solution of cost 1.28 between R-NN heuristic and K-NN.

### 5.3 Effective use of R-NN in MP algorithms in practice

We now discuss how to effectively use the radial connection scheme of asymptotically (near) optimal MP algorithms. We first describe a common scenario for which the computed radii are practically useless, and continue by suggesting a simple heuristic to overcome this problem.

The proofs of asymptotic optimality provided by Karaman and Frazzoli [28] and by Janson et al. [27] rely on the following implicit assumption:

**Assumption 5** *For $x \in \mathcal{X}$, w.h.p. $\mathbb{B}_\mathcal{X}(r, x) = \mu(\mathcal{B}_\mathcal{X}(r, x) \cap \mathcal{X})$.*

This assumption does not hold when the center of the ball is close to the boundary of the C-space $\mathcal{X}$. The reason why this assumption may be used in the aforementioned proofs is that the proofs consider balls of radii proportional to $r_n$ which tends to zero as $n \to \infty$. This, in turn, implies that, for a given path $\sigma$, there exits a certain number of samples $n_0$ for which the radius $r_{n_0}$ is sufficiently small and hence the sequence of balls of radius $r_{n_0}$ covering $\sigma$ does not intersect the boundary of $\mathcal{X}$.

In many common settings, Assumption 5 does not hold for practical values of $n$. Consider, for example, Fig. 9a, which depicts a two-dimensional rectangular C-space where one dimension is significantly larger than the other. For small values of $n$, any ball of radius $r_n$ intersects the boundary of $\mathcal{X}$ (as the ball $\mathcal{B}(r_n, x)$, drawn in orange, for which $\mathcal{B}(r_n, x) \setminus \mathcal{X} \neq \emptyset$). As a result, the number of configurations of distance at most $r_n$ from a configuration $x$ might be too small, and this, in turn, may cause the roadmap of $n$ vertices to remain disconnected.

We start by formally describing the setting and propose a heuristic to choose a larger radius. Let $\mathcal{X} = \mathcal{X}_1 \times \mathcal{X}_2$ be a $d$-dimensional compound C-space and assume that $\mu(\mathcal{X}_1) \geq \mu(\mathcal{X}_2)$. Let $d_1, d_2$ denote the dimensions of $\mathcal{X}_1, \mathcal{X}_2$, respectively. Finally, let $\rho_{\max(\mathcal{X}_2)}$ be the maximal distance between any two points in $\mathcal{X}_2$ and assume that $\rho = w_1\rho_1 + w_2\rho_2$. When Assumption 5 does not hold, as in Fig. 9, the intuition is that the "effective dimension" of our C-space is closer to $d_1$ than to $d_1 + d_2$. If $r_n > w_2\rho_{\max(\mathcal{X}_2)}$ then $\forall x \in \mathcal{X} \ \mathbb{B}_\mathcal{X}(r_n, x) > \mu(\mathcal{B}_\mathcal{X}(r_n, x) \cap \mathcal{X})$. In such cases, we suggest to project all points to $\mathcal{X}_1$ and use the critical connection radius that we would have used had the planning occurred in $\mathcal{X}_1$.

To evaluate the proposed heuristic, we ran Lazy-sPRM* on the Cubicles scenario (Fig. 2c) using R-NN with and without the heuristic, and also using K-NN strategy (with the standard $k_n$ value). We measured the cost of the solution path as a function of the running time. As depicted in Fig. 9b, the heuristic was able to find higher-quality solutions using less samples, resulting in a tenfold speedup in obtaining a solution of a certain cost, when compared to K-NN. Moreover, R-NN without the heuristic was practically inferior, as it was not able to find a solution even for large values of $n$; results omitted.

## 6  Discussion and future work

We have described settings where the computational role of NN search after a finite time is far from negligible when compared to that of CD. We have characterized these settings and demonstrated this phenomenon empirically. By developing improved NN-search techniques that are tailored for MP, the overall running time of planners in NN-sensitive settings can be significantly reduced.

Identifying NN-sensitive settings is not always straightforward. Especially in the case of NN-sensitive parameters, where one needs to determine whether a set of parameters is sub-optimal for the problem at hand. It would be interesting to identify additional NN-sensitive parameters, perhaps algorithm-specific ones that are less general. Another open problem is whether it is possible to develop a parameter-tuning technique (or a heuristic) such that the computational weight of NN is increased. Then, by using NN techniques tailored to MP, the overall performance of the algorithm can be significantly improved.

## A  Volume of ball in common C-spaces

In this section we derive closed-form expressions for the volume of the unit ball in many common C-spaces and distance metrics. We start with the $d$-dimensional Euclidean space $\mathbb{R}^d$ (considering two different metrics) and continue with the $d$-dimensional unit sphere embedded in $\mathbb{R}^{d+1}$ with geodesic distance as the metric. These results are summarized in Table 6. We then continue to consider a set of compound spaces for which the metric is defined by setting $p = 1$ in Eq. 4. This choice follows the way metrics are defined in OMPL [15] for compound spaces. All results can be extended straightforwardly to different values of $p$.

$L^2$—*Euclidean space with Euclidean metric*  Arguably, the most simple space to be considered is the Euclidean space. Here $\mathcal{X} = \mathbb{R}^d$, and given two points $x, y \in \mathcal{X}$ such that $x = (x_1, \ldots, x_d)$, $x_i \in \mathbb{R}$ (similarly for $y$), the distance metric is the standard Euclidean metric. Namely, $\rho(x, y) = \left(\sum_i (x_i - y_i)^2\right)^{1/2}$. We denote by $L^2(d)$ the $d$-dimensional Euclidean space where $\rho$ is the Euclidean metric.

Computing the volume of the unit ball in this space is a well studied problem. For a detailed discussion on properties of the unit ball see `https://en.wikipedia.org/wiki/Volume_of_an_n-ball`.

We have that,

$$\mathbb{B}_{L^2(2d)}(r) = \frac{\pi^d}{d!} r^{2d}; \quad \mathbb{B}_{L^2(2d+1)}(r) = \frac{2(d!)(4\pi)^d}{(2d+1)!} r^{2d+1}; \quad \mathbb{S}_{L^2(d)}(r) = \frac{d}{r} \mathbb{B}_{d,r}.$$

It is straightforward to see that $L^2(d)$ is a well-behaved space (according to Definition 3).

$L^1$—*Euclidean space with Manhattan metric*  We consider the Euclidean space $\mathcal{X} = \mathbb{R}^d$ with a different distance metric, namely the Manhattan metric. Here $\rho(x, y) = \sum_i |x_i - y_i|$. This is less common in motion-planning applications, but serves as a simple example to several concepts that will be used. We denote by $L^1(d)$ the $d$-dimensional Euclidean space where $\rho$ is the Manhattan metric.

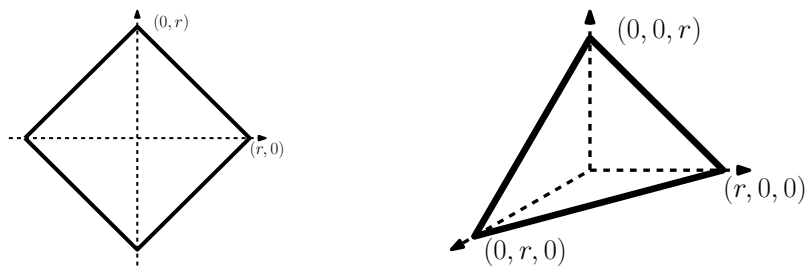

**Fig. 10:** Balls of radii $r$ in $L^1$. Two dimensional ball (left) and one octant of a three dimensional ball (right).

Computing the volume of the unit ball in this space is also well-studied (see, e.g., [52]). Here we have that $\mathbb{B}_{L^1(d)}(r) = \left(2^d/d!\right) \cdot r^d$. Of specific interest are the two- and three-dimensional cases where (see Fig. 10),

$$\mathbb{B}_{L^1(2)}(r) = 2r^2; \quad \mathbb{B}_{L^1(3)}(r) = \frac{4}{3}r^3; \quad \mathbb{S}_{L^1(2)}(r) = 4\sqrt{2}r; \quad \mathbb{S}_{L^1(3)}(r) = 2\sqrt{3}r^2.$$

Note that this C-space is not well behaved. Consider for example the two-dimensional case. Clearly,

$$\frac{\partial \mathbb{B}_{L^1(2)}(r)}{\partial r} = \frac{\partial 2r^2}{\partial r} = 4r \neq 4\sqrt{2}r = \mathbb{S}_{L^1(2)}(r).$$

However, as described in [40], for any $d$-dimensional C-space $\mathcal{X}$ the transformation $s(r) = d \cdot \mathbb{B}_{\mathcal{X}}(r)/\mathbb{S}_{\mathcal{X}}(r)$ can be applied. Expressing the volume of a ball and a sphere using $s(r)$, instead of $r$, yields representations for which the ball-sphere derivative property holds. We call these representations *canonical* and denote them by $\tilde{\mathbb{B}}_{\mathcal{X}}(s(r))$ and $\tilde{\mathbb{S}}_{\mathcal{X}}(s(r))$, respectively. Lemma 4 can then be used even if a space is not well behaved, assuming that one can produce analytic representations of $\tilde{\mathbb{B}}_{\mathcal{X}}(s(r))$ and $\tilde{\mathbb{S}}_{\mathcal{X}}(s(r))$.

$S^d$—*Unit Sphere with geodesic distance* Let $\mathcal{X}$ be the unit hyper-sphere $S^d$ embedded in $\mathbb{R}^{d+1}$, i.e., $\mathcal{X} = S^d \equiv \{q \in \mathbb{R}^{d+1} \mid \sum_i x_i^2 = 1\}$. We use geodesic distance as our metric: $\rho(q_1, q_2) = \arccos |q_1 \cdot q_2|$. We study this space as it will allow us to obtain results for the spaces representing rotations in two and three dimensions. It is well known that $S^1 \cong SO(2)$ where $A \cong B$ denotes that $A$ is homeomorphic to $B$. When considered as the set of unit quaternions with absolute value 1, $S^3 \simeq SO(3)$ where $A \simeq B$ denotes that $A$ is diffeomorphic to $B$. See, e.g., [36, Chapter 4] for an in-depth discussion on different topological structures and their connection to motion-planning configuration spaces. Choosing geodesic distances is a common choice to compare rotations, see [23] for a comparative study on different distance metrics for rotations. Furthermore, the $d$-dimensional torus $T(d)$ is defined as the Cartesian product of $d$ unit circles namely, $T(d) = \left(S^1\right)^d$. This is of interest as the configuration space of a planar articulated robot with $m$ links is $R^2 \times T(m)$. Thus, we concentrate on $S^d$ for $d = 1, 2$ and 3.

Li [37] gives a general formula for the area of a hyper-spherical cap which is exactly $\mathbb{B}_{S^d}(r)$ in our setting. Specifically, we have that

$$\mathbb{B}_{S^1}(r) = 2r; \quad \mathbb{B}_{S^2}(r) = 2\pi(1 - \cos r); \quad \mathbb{B}_{S^3}(r) = \pi \left(2r - \sin(2r)\right).$$

To compute $\mathbb{S}_{S^d}(r)$, note that the intersection of a hyperplane perpendicular to one of the axes with a $d$-dimensional hyper-sphere is a hyper-ball of dimension $d - 1$. The radius of this ball is exactly $\sin r$ and we need its surface area (see Fig. 11). Thus, we have that

$$\mathbb{S}_{S^1}(r) = 2; \quad \mathbb{S}_{S^2}(r) = 2\pi \sin r; \quad \mathbb{S}_{S^3}(r) = 4\pi \sin^2 r.$$

Note that $S^1$ is well behaved while $S^2$ and $S^3$ are not. Using $s(r) = 2(1 - \cos r)/\sin r = 2 \tan \frac{r}{2}$ one can obtain the canonical forms for $S^2$. Similarly, using $s(r) = 3(2r - \sin(2r))/(4 \sin^2 r)$ the canonical forms for $S^3$ can be obtained.

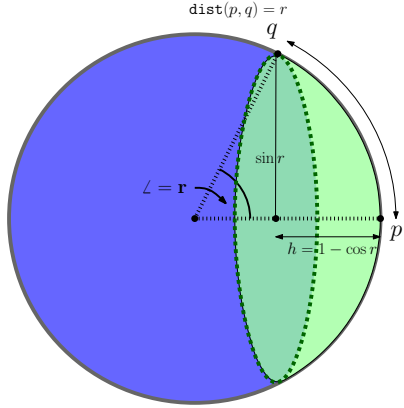

**Fig. 11:** Visualization of a spherical cap on the unit sphere $S^2$ embedded in $\mathbb{R}^3$.

*SE(2)—The configuration space of a planar translating and rotating rigid body*
Recall that $SE(2) = L^2(2) \times S^1$, the volume of a sphere in $L^2(2)$ is $2\pi r$, while the volume of a ball in $S^1$ is $2r$. In addition, recall that $L^2(2)$ is well behaved. Using Eq. 4 as our distance metric and using Lemma 4 we have

$$\mathbb{B}_{SE(2)}(r) = \int_{\varrho \in [0, r/w_1]} (2\pi\varrho)(2(r - w_1\varrho)/w_2)d\varrho = \frac{2\pi}{3}\frac{1}{w_1^2 w_2}r^3. \qquad (7)$$

*T(d)—The configuration space of a kinematic chain in the plane* Recall that $T(1) = S^1$ and that $T(d) = S^1 \times T(d-1)$. Furthermore, the volume of a sphere in $S^1$ is 2 while the volume of a ball in $S^1$ is $2r$. In addition, recall that $S^1$ is well behaved. We start with the simple case of a two-dimensional torus $T(2)$. Here, using Eq. 4 as our distance metric and using Lemma 4 we have

$$\mathbb{B}_{T(2)}(r) = \int_{\varrho \in [0, r/w_1]} 2 \cdot 2\frac{r - w_1\varrho}{w_2}d\varrho = \frac{2}{w_1 w_2}r^2. \qquad (8)$$

| C-space | $L^2(2)$ | $L^2(3)$ | $L^1(2)$ | $L^1(3)$ | $S^1$ | $S^3$ |
|---|---|---|---|---|---|---|
| $\mathbb{B}(r)$ | $\pi r^2$ | $\frac{4}{3}\pi r^3$ | $2r^2$ | $\frac{4}{3}r^3$ | $2r$ | $\pi(2r - \sin(2r))$ |
| $\mathbb{S}(r)$ | $2\pi r$ | $4\pi r^2$ | $4\sqrt{2}r$ | $2\sqrt{3}r^2$ | $2$ | $4\pi \sin^2 r$ |
| $\mathbb{B}(1)$ | $\pi$ | $\frac{4}{3}\pi$ | $2$ | $\frac{4}{3}$ | $2$ | $\pi(2 - \sin(2))$ |

**Table 6:** Summary of $\mathbb{B}(r), \mathbb{S}(r)$ and $\mathbb{B}(1)$ for several C-spaces.

In general, the volume of a ball in T($d$) using the distance metric $\rho_{T(d)} = \sum_i w_i \rho_{S^1}$ is computed by using the recursive definition of T($n$) to obtain

$$\mathbb{B}_{T(d)}(r) = \left( \frac{2^d}{d!} \prod_i \frac{1}{w_i} \right) \cdot r^d. \tag{9}$$

$L^2(d) \times L^2(d)$—*The configuration space of two translating robots* Considering the Cartesian product of two Euclidean spaces comes naturally in the case of two translating robots in $\mathbb{R}^d$ for $d \in \{2, 3\}$. Recall that $\forall d$ $L^2(d)$ is well behaved. Thus, for the case of $d = 2$ and using Eq. 4 as our metric we have

$$\mathbb{B}_{L^2(2) \times L^2(2)}(r) = \int_{\varrho \in [0, r/w_1]} (2\pi \varrho) \cdot \pi \left( \frac{r - w_1 \varrho}{w_2} \right)^2 d\varrho = \frac{\pi^2}{6} \frac{1}{w_1^2 w_2^2} r^4. \tag{10}$$

For the case of $d = 3$ and using Eq. 4 as our metric we have

$$\mathbb{B}_{L^2(3) \times L^2(3)}(r) = \int_{\varrho \in [0, r/w_1]} (4\pi \varrho^2) \cdot \frac{4\pi}{3} \left( \frac{r - w_1 \varrho}{w_2} \right)^3 d\varrho = \frac{4\pi^2}{45} \frac{1}{w_1^3 w_2^3} r^6. \tag{11}$$

$L^2(2)^m$—*The configuration space of $m$ translating planar robots* We generalize the above results for the case of $m$ translating planar robots. The metric considered is $\rho_{(L^2(2))^m} = \Sigma_{i=1}^m w_i \rho_{L^2(2)}$.

By generalizing the above approach, we have that

$$\mathbb{B}_{(L^2(2))^m} = C_m \pi^m \cdot \left( \prod_{i=1}^m \frac{1}{w_i^2} \right) \cdot r^{2m}, \tag{12}$$

where $C_m = \frac{C_{m-1}}{m(2m-1)}$ and $C_1 = 1$.

$L^2(3)^m$—*The configuration space of $m$ translating spatial robots* For the case of $m$ translating spatial robots, when the metric considered is $\rho_{(L^2(3))^m} = \Sigma_{i=1}^m w_i \rho_{L^2(3)}$ and using similar arguments, we have that

$$\mathbb{B}_{(L^2(3))^m} = C_m \pi^m \cdot \left( \prod_{i=1}^m \frac{1}{w_i^3} \right) \cdot r^{3m}, \tag{13}$$

Where $C_m = \frac{8 C_{m-1}}{3m(3m-1)(3m-2)}$ and $C_1 = 4/3$.

$(SE(2))^m$—*The configuration space of $m$ translating and rotating planar robots* We generalize the result we obtained for two robots for the case of $m$ translating and rotating planar robots. The metric considered is $\rho_{(SE(2))^m} = \Sigma_{i=1}^m w_{2i-1} \rho_{L^2(2)} + w_{2i} \rho_{S^1}$.

We have that

$$\mathbb{B}_{(SE(2))^m} = C_m \pi^m \cdot \left( \prod_{i=1}^m \frac{1}{w_{2i-1}^2 w_{2i}} \right) \cdot r^{3m}, \tag{14}$$

where $C_m = \frac{4 C_{m-1}}{3m(3m-1)(3m-2)}$ and $C_1 = 2/3$.